\documentclass{article} % For LaTeX2e

\usepackage{amsmath}
\usepackage{amssymb}
\PassOptionsToPackage{numbers,compress}{natbib}
\usepackage[preprint]{neurips_2026}
\workshoptitle{Continual Learning in the Era of Foundation Models and Embodied Agents (CL4FMAgents), NeurIPS 2026}

\usepackage{amsmath,amsfonts,bm}

\def\eqref#1{equation~\ref{#1}}
\def\1{\bm{1}}

\DeclareMathAlphabet{\mathsfit}{\encodingdefault}{\sfdefault}{m}{sl}
\SetMathAlphabet{\mathsfit}{bold}{\encodingdefault}{\sfdefault}{bx}{n}

\usepackage{hyperref}
\usepackage{url}
\usepackage{booktabs}
\usepackage{array}
\usepackage{graphicx}
\usepackage{multirow}
\usepackage{wrapfig}
\usepackage{xcolor}
\IfFileExists{ulem.sty}{\usepackage[normalem]{ulem}}{}

\newcommand{\parpct}{\ensuremath{\mathrm{PR}\%}}
\newcommand{\passat}[1]{pass@#1}
\definecolor{jeancolor}{HTML}{B8860B}

\title{CHART: A Harness-Rotation Curriculum for\\ Harness-Robust Search Agents}

\author{%
  Xinlu Zhang\thanks{Correspondence: \texttt{zhaxinlu@amazon.com}}
  \quad Ying-Chun Lin
  \quad Zhihan Zhang
  \quad Besnik Fetahu
  \quad Xi Chen\\
  Amazon\\
  \texttt{zhaxinlu@amazon.com}
}

\begin{document}

\maketitle

\begin{abstract}
Search agents are usually trained under a single harness. But once an agent is deployed in a real application, its harness is frequently updated (e.g., a rewritten system prompt) to fit production needs. This exposes a fragility of post-trained agents: because a learned behavior is entangled with its training harness, even a harness update that leaves the task unchanged can fail to elicit the behavior. In this study, we train a search agent to perform parallel search, a popular strategy for improving both the efficiency and the performance of search. We find that training under a fixed harness makes the behavior \emph{harness-local}, overfit to that harness's surface form: when the harness changes, the model falls back to serial search. An intuitive idea is \emph{harness augmentation}, but simply creating more harnesses for training does not completely resolve the problem. GRPO learns from the reward gap between parallel and serial rollouts of the same question: a small harness pool saturates that gap early, while a large pool dilutes the per-harness signal too thinly for any harness to consolidate. We therefore propose Curriculum HArness Rotation Training (\textbf{CHART}), a rotating curriculum that lets a search agent gradually consolidate parallel search across harnesses. At each periodic evaluation, CHART ``graduates'' the harnesses on which the expected behavior has been learned and replaces them with harnesses that are still learnable, which keeps the reward gap alive throughout training. Starting from the same harness pool, CHART makes the model learn parallel search on all harnesses, whereas static augmentation succeeds on at most half of them. The behavior also carries to held-out harnesses: CHART parallelizes on $89\%$ of held-out turns, against at most $5\%$ for the static pools. It further transfers to a new QA task and search environment, improving \passat{1} by $5.6$pp over the best static pool. Besides, we show that CHART-trained agents benefit more from meta-harness search than other baselines.

\end{abstract}

% ============ INTRODUCTION ============
\section{Introduction}
\label{sec:intro}

Language-model agents increasingly act by calling external tools---searching the
web, querying APIs, running code---inside a multi-turn loop. Reinforcement
learning (RL) with verifiable rewards is now the standard way to teach these
agents useful behaviors \citep{jin2025searchr1,song2025r1searcher}, one of the
most valuable being \emph{intra-turn parallelism}: issuing several search queries
at once rather than one at a time, which cuts latency and, on many questions,
improves coverage \citep{kim2023llmcompiler}.\footnote{Appendix~\ref{app:parvsseq}
verifies both in our setting with a budget-matched sequential baseline.} But an agent is
never deployed as bare weights. It runs behind a \emph{harness}---a system prompt
that frames the task, names the tools, and implicitly or explicitly suggests a
search strategy---and this harness is rewritten constantly downstream: a product
team adds a persona, translates it, imposes a format, or strips it to one
line. Behaviors acquired through RL should be \emph{robust} to these rewrites. We show
they are not.

\paragraph{Learned behavior is harness-local.}
Under a fixed training harness, RL installs parallel search as a habit tied to that
harness's surface form: a model that parallelizes on $91\%$ of turns under its own
training harness holds only $45\%$ on harnesses that still ask for parallel search, and
just $2\%$ on those that do not (\S\ref{sec:motivation}). The capability is intact---the model issues parallel
searches only
when the wording matches training, the agentic analogue of the prompt sensitivity of
frozen LLMs \citep{sclar2023quantifying,lu2022fantastically,zhu2023promptrobust}. A
routine deployment-time rewrite thus silently deletes a behavior the model provably
possesses.

\paragraph{Static harness augmentation does not fix it.} The obvious response is
\emph{harness augmentation}---train over many harness variants so that
no single phrasing is load-bearing, the instinct behind multi-template instruction
tuning \citep{wei2021flan,sanh2021t0,sun2023robustness} and, now, frontier agent
training \citep{kimiteam2026k3}. Yet \emph{static} mixing fails at every pool size:
for both small and large training pools, parallel search becomes the default
on only a minority of harness styles (\S\ref{sec:motivation})---more variants buy
\emph{cue-following}, not \emph{internalization}. The failure is in the RL
dynamics, not the pool size: GRPO \citep{shao2024deepseekmath} learns from a
\emph{within-group contrast}---the reward gap between parallel and serial rollouts
of the same question---which a small pool exhausts (its harnesses saturate and the
contrast vanishes) while a large pool dilutes, splitting each step's rollout budget so
thinly across harnesses that none consolidates. Static augmentation is trapped between
the two.

\paragraph{Curriculum harness augmentation.} The fix is to \emph{schedule} the
harnesses rather than mix them, so the within-group contrast never runs out.
\textbf{CHART}, Curriculum HArness Rotation Training, keeps a small active window of
harnesses; the moment one saturates, CHART ``graduates'' it and swaps in a fresh,
still-learnable one, so every batch keeps a live parallel-vs-serial contrast
and the behavior spreads across structurally diverse styles. This is a curriculum
over the \emph{harness}, not the problem: prior curriculum RL orders \emph{problems}
by difficulty \citep{bengio2009curriculum,xi2024reverse,jiang2025vcrl,qi2024webrl},
whereas we order which harness styles are active over training. Critically, rotation
draws from the \emph{same} pool as static augmentation; only its organization in
time changes.

\paragraph{Findings.} From the identical pool, CHART makes parallel search the
default on $100\%$ of harnesses (up from at most $48\%$ under static augmentation),
generalizing to cue-free and \emph{never-trained} held-out styles and beating every
static pool by $5.6$pp \passat{1} on a new QA task and search environment, at no
accuracy cost. The through-line is that \emph{sequencing matters more than
diversity}. Robustness also pays off downstream: a CHART-trained model has a
\emph{flatter harness-loss landscape}, making meta-harness search
\citep{yang2023opro,zhou2022ape,agrawal2025gepa} converge to better harnesses,
$+12.6$pp over an untrained substrate.

\paragraph{Contributions.}
\begin{itemize}\itemsep2pt
  \item \textbf{A new failure mode for agentic RL.} RL under a fixed harness makes a
    learned behavior \emph{harness-local}---the model keeps the capability but
    deploys it only under training-style prompts---and \emph{static} harness
    augmentation fails at \emph{every} pool size, exhausting the within-group
    contrast RL learns from (\S\ref{sec:method}, \S\ref{sec:motivation}).
  \item \textbf{Curriculum harness augmentation.} We introduce, to our knowledge,
    the first curriculum over the \emph{harness interface} rather than problem
    difficulty (\textbf{CHART}): from the \emph{same} pool it turns harness-local
    behavior into a default robust across unseen styles, never-trained held-out
    harnesses, and an out-of-distribution (OOD) task and environment, at no accuracy
    cost (\S\ref{sec:method}, \S\ref{sec:main}, \S\ref{sec:ablation}).
  \item \textbf{Robustness as a better substrate.} Harness-robustness reshapes the
    model's harness-loss landscape, making a CHART-trained model a strictly better
    substrate for automatic meta-harness search ($+12.6$pp, \S\ref{sec:metasearch}).
\end{itemize}

% ============ RELATED WORK ============
\section{Related Work}
\label{sec:related}

\paragraph{Harness robustness.} Frozen LLMs are notoriously sensitive to superficial
prompt changes---formatting, ordering, paraphrasing, and adversarial perturbation all
shift accuracy, and a single prompt overstates ability
\citep{sclar2023quantifying,lu2022fantastically,young2025follow,brittlebench2026,zhu2023promptrobust,mizrahi2024state}.
These works
\emph{measure} prompt sensitivity; we \emph{remove} it during RL, and unlike
inference-time hardening the robustness \emph{generalizes to unseen harness styles}.
Our failure mode is its agentic analogue: the same task, reworded, flips the agent
between parallel and serial search.

\paragraph{RL for search and tool-using agents.} RL with verifiable rewards teaches
LLMs to search and use tools---Search-R1 \citep{jin2025searchr1}, R1-Searcher
\citep{song2025r1searcher}, ZeroSearch \citep{sun2025zerosearch}, ToRL
\citep{li2025torl}, and long-horizon search agents \citep{kalyan2025longhorizon}---but
targets \emph{acquiring} competence under a single fixed prompt. We study an
orthogonal axis, the robustness of a learned behavior to the harness, and show
fixed-prompt training produces harness-local behavior that fails to generalize.
Concurrent frontier work varies the harness during RL---Kimi~K3 trains over
composable harness configurations to avoid overfitting to any single tool schema or
system prompt \citep{kimiteam2026k3}---but mixes them statically; we ask whether the
learned behavior stays harness-local and show that \emph{scheduling}, not mixing,
internalizes it.

\paragraph{Curriculum RL for LLMs.} Curriculum learning \citep{bengio2009curriculum}
for LLM RL orders \emph{problems} by difficulty \citep{xi2024reverse,jiang2025vcrl,qi2024webrl}
or synthesizes problems targeting model weaknesses \citep{liang2025sws}; distributionally
robust objectives push for uniform performance across groups \citep{panaganti2026gdro}.
Our rotation is instead a curriculum over the \emph{harness}, aiming at behavioral
generalization across harnesses rather than accuracy on hard problems. The
weakness-driven selection of \citet{liang2025sws} resembles our swapping in
low-parallel harnesses, but acts on synthesized problems, not the harness interface.

\paragraph{Automatic prompt and harness optimization.} A large line of work searches
for a better prompt with the model fixed---OPRO, APE, evolutionary search, and agentic
optimizers
\citep{yang2023opro,zhou2022ape,guo2023evoprompt,zhang2024revisiting,maspromptbench2026,agrawal2025gepa}---and,
closest to us, Meta-Harness \citep{lee2026metaharness} optimizes the harness
end-to-end. Our meta-harness
search (\S\ref{sec:metasearch}) is in this lineage but serves as a \emph{probe}:
CHART changes the \emph{weights} to be robust across harnesses, and we ask whether
that makes harness search easier.

% ============ METHOD ============
\section{Problem and Method}
\label{sec:method}

\subsection{Notation}
We train a search agent with group-relative policy optimization (GRPO)
\citep{shao2024deepseekmath}. The agent answers each question in at most $T$ turns,
each issuing up to $P$ searches at once before reading their results: a $T{\times}P$
search budget in which $T$ is \emph{depth} (serial multi-hop search) and $P$ is
\emph{width} (parallel search). A turn is \emph{parallel} if it issues $>1$ call. A
\emph{harness} $h$ is a system prompt framing the task; the same question can appear
under many harnesses that differ in surface form while leaving the task unchanged
(\S\ref{sec:setup}). Its \emph{parallel rate} \parpct{} is the fraction of turns that
are parallel, over eval rollouts under $h$, and $h$ is \emph{covered} when
$\parpct\ge0.80$. We report both mean \parpct{} and \emph{coverage}, the number of
covered harnesses: mean \parpct{} can be lifted by partial parallelism spread
everywhere, whereas coverage counts where parallel search is reliably the default.

\paragraph{Reward.} The reward is purely outcome-based:
$r = r_{\mathrm{correct}} + 0.1\,r_{\mathrm{format}}$, an LLM-judge verdict
$r_{\mathrm{correct}}\in\{0,1\}$ on the final answer plus a format check. \emph{No term
rewards parallelism}, so any parallel-vs-serial preference arises entirely from the
within-group correctness contrast (\S\ref{sec:motivation}).

\subsection{CHART: Curriculum HArness Rotation Training}
\label{sec:chart}
CHART trains the same pool as a \emph{curriculum over the harness} rather than a
static mixture. GRPO learns from a \emph{within-group contrast}---the reward gap
between parallel and serial rollouts of the same question---which any fixed pool
eventually exhausts (\S\ref{sec:motivation}). CHART instead keeps only a small
\emph{rotating window} of active harnesses, \emph{graduating} each as it saturates
and \emph{replacing} it with a fresh, still-learnable one; because saturated
harnesses are continually swapped out, the window always retains a live contrast, and
parallel search spreads across the whole pool.

\paragraph{The rotating window.} From a pool $\mathcal{H}$ ($|\mathcal{H}|{=}33$) we
keep $N$ \emph{active} harnesses $A_t\subseteq\mathcal{H}$ at eval step $t$. Write
$\parpct_t(h)\in[0,1]$ for the parallel rate of harness $h$ on the step-$t$ eval and
$\mathcal{G}_t$ for the graduated set. Training runs as usual on $A_t$; the two rules
below say who leaves and who arrives.

\paragraph{Graduation trigger.} Every 20 steps we run a lightweight in-pool eval
(50 datapoints per harness) to estimate $\parpct_t(h)$ for each active harness, and
graduate every active harness that eval shows as covered:\footnote{We graduate at
$0.85$ rather than at the $0.80$ coverage bar itself---a margin that absorbs the noise
of a 50-datapoint estimate (Appendix~\ref{app:evalsize}) and keeps arrival strictly
below departure.}
\begin{equation}
  \mathcal{G}_t = \mathcal{G}_{t-1}\cup\{\,h\in A_{t-1}: \parpct_t(h)\ge0.85\,\}.
\end{equation}
Because graduation is checked only at these periodic evals, a harness keeps
training past the bar until the next eval, which we find matters for consolidation
(\S\ref{sec:ablation}).

\paragraph{Replacement selection.} We refill the freed slots from the learnable,
not-yet-graduated candidates $\mathcal{H}\setminus(\mathcal{G}_t\cup A_{t-1})$,
choosing those whose parallel rate is closest to the center $c{=}0.60$ of a band
$[0.40,0.80]$ whose upper end is the coverage bar itself, so the incoming harness is
learnable but not yet covered---closer to graduation and it has little left to teach:
\begin{equation}
  A_t = \bigl(A_{t-1}\setminus\mathcal{G}_t\bigr)\;\cup\;
  \operatorname*{arg\,min}_{\,S}\;\sum_{h\in S}\bigl|\parpct_t(h)-c\bigr|,
  \quad
  S\subseteq \mathcal{H}\setminus(\mathcal{G}_t\cup A_{t-1}),\;
  |S| = N-|A_{t-1}\setminus\mathcal{G}_t|.
\end{equation}
When $N{>}1$ the swap is \emph{staggered}---only graduated harnesses are replaced, so
the surviving ``anchors'' stabilize the batch across the switch. If no candidate lies
in the band we take the highest-\parpct{} one, and once the pool is exhausted the
survivors keep training to convergence.

% \paragraph{Turn schedule.} We optionally warm up under a capped turn budget
% (fewer turns) before opening the full budget. Capping turns removes the serial
% depth advantage during the fragile early phase, so parallel is the only path to
% higher reward; the full budget is later opened for depth learning. Rotation and
% turn schedule are orthogonal knobs; \S\ref{sec:ablation} isolates each.

% ============ EXPERIMENTAL SETUP ============
\section{Experimental Setup}
\label{sec:setup}

\paragraph{Data and model training.} We train Qwen3-30B-A3B-Thinking-2507
\citep{qwen3} with GRPO (global batch 512, $16$ samples $\times$ $32$ groups) on
$32$ H100 GPUs. Our data is MiroVerse-v0.1 \citep{miroverse2025}, keeping only
questions answerable with web search. A difficulty filter runs the base model $16$
times per question, scores each with an LLM judge, and uses the \emph{effective
pass rate} (EPR, fraction correct) to split questions into \emph{easy}
($\mathrm{EPR}>0.60$), \emph{learnable} ($0<\mathrm{EPR}\le0.60$), and \emph{hard}
($\mathrm{EPR}=0$). We train on the learnable band ($16{,}712$ questions) and sample
$50$ per band for the $150$-question val set used in \S\ref{sec:motivation},
\S\ref{sec:main}, and \S\ref{sec:ablation}. Each run trains for 200 steps at 512
rollouts/step with $T{=}5$ turns and $P{=}5$ calls per turn, a $25$-search cap per
question.

\paragraph{Harness pool.} We author 33 harnesses with Claude Code \citep{anthropic2025claudecode} spanning
wide surface diversity---wording, persona, language, verbosity,
and agent format---grouped by \emph{how much the harness tells the model what to do}
(Appendix~\ref{app:harnesses}): \textbf{A--Explicit} (10 harnesses) state the parallel strategy
outright; \textbf{B--Implicit} (9) only hint at it; \textbf{C--Framework} (9) recast
the task in agent formats (ReAct, plan-and-execute, function/XML calls); and
\textbf{D--Minimal} (5) give near-zero cues, testing whether parallel search is
internalized. A/C are \emph{cue} harnesses that request or frame parallel search; B/D are
\emph{cue-free}, never mentioning it---the split between ``parallelizes only when
asked'' and ``parallel-as-default'' used throughout. Each run samples one harness per
GRPO group; the methods we compare differ in which they draw (\S\ref{sec:motivation}).

\paragraph{Evaluation protocol.}
\begin{wraptable}{r}{0.60\linewidth}
\vspace{-1.4em}
\centering
\small
\setlength{\tabcolsep}{4pt}
\caption{Eval tasks; each reports \passat{1} and \parpct{}.}
\label{tab:eval-protocol}
\begin{tabular}{@{}lcl@{}}
\toprule
task & harn.$\times$data & role \\
\midrule
in-pool  & $33{\times}150$ & all 33 harnesses, incl.\ untrained \\
held-out & $12{\times}150$ & 12 unseen harnesses \\
shopping & $19{\times}200$ & new task $+$ search env \\
\bottomrule
\end{tabular}
\end{wraptable}
We evaluate on three tasks of increasing distance
from training (Table~\ref{tab:eval-protocol}). \textbf{In-pool robustness}
($33$ harnesses $\times$ $150$ questions) scores every model on all 33 pool harnesses, including untrained ones;
\textbf{held-out harnesses} ($12\times150$) reuses the same \texttt{miroverse}
questions with $12$ harnesses in no training pool---our primary cross-harness
\emph{robustness} metric. None of the 12 asks for parallel search: each states the
per-turn call budget and otherwise leaves the strategy open, so the held-out task is
entirely cue-free. Finally, the \textbf{shopping agent task} ($19\times200$) swaps
the data itself, a separate product-search set (its validation split;
\S\ref{sec:metasearch} uses the test split) and search environment, testing
transfer to a new QA distribution. The 12 held-out and 19 shopping harnesses were
authored with Claude Code \citep{anthropic2025claudecode} like the training pool, but no evaluated style overlaps a
trained one. All evals use the 5-turn/5-parallel environment
and \passat{1} at temperature 1.0. We select checkpoint by best
average \passat{1} over the 33 pool harnesses,\footnote{Selection and graduation read
a lighter $33\times50$ in-pool eval---selection on average \passat{1}, graduation on
each harness's \parpct{}---agreeing with the full $33\times150$ eval on 97\% of
graduation decisions (Appendix~\ref{app:evalsize}).} then run all three tasks
\emph{once} on it.
% ============ RESULTS ============
\section{Results}
\label{sec:results}

We conduct one controlled sweep over two axes---pool \emph{size}, and whether the pool
\emph{rotates}---at a matched 200-step budget: each fixed-$N$/CHART-$N$ pair is seeded
from the same $N$ harnesses and differs \emph{only} in whether the window moves. We
report the in-pool results in \S\ref{sec:motivation} and the held-out and out-of-domain
results in \S\ref{sec:main}, probe the trained models with meta-harness search in
\S\ref{sec:metasearch}, and ablate the design in \S\ref{sec:ablation}.

\begin{table}[t]
\centering
\small
\setlength{\tabcolsep}{4pt}
\caption{All models at 200 training steps on the three validation tasks of
\S\ref{sec:setup}; no model trains on the held-out or shopping harnesses. \parpct{} is
a per-harness mean, split in-pool into \emph{cue} and \emph{cue-free}
(\S\ref{sec:setup}), with \emph{mean} over all 33. \emph{cov.}\ is coverage
($\parpct\ge0.80$, \S\ref{sec:method}). CHART stops at $N{=}9$ since \mbox{fixed-33} is
also \mbox{CHART-33}. Bold and underline mark the best and second-best \passat{1}; the
\parpct{} columns are read as the cue-vs-cue-free gap, not as a maximum.}
\label{tab:main}
\begin{tabular*}{\linewidth}{@{\extracolsep{\fill}}l ccccc cc cc@{}}
\toprule
& \multicolumn{5}{c}{in-pool (33 harnesses)} & \multicolumn{2}{c}{held-out harnesses} & \multicolumn{2}{c}{shopping (OOD)} \\
\cmidrule(lr){2-6}\cmidrule(lr){7-8}\cmidrule(lr){9-10}
& & \multicolumn{3}{c}{\parpct} & & & & & \\
\cmidrule(lr){3-5}
Model & \passat{1} & cue & cue-free & mean & cov. & \passat{1} & \parpct & \passat{1} & \parpct \\
\midrule
base (no train) & 0.364 & 37\% & 5\% & 24\% & 3/33 & 0.358 & 2\% & 0.375 & 5\% \\
\midrule
\multicolumn{10}{@{}l}{\emph{fixed pool of $N$ harnesses}} \\
fixed-1    & 0.456 & 45\% & 2\% & 27\% & 6/33 & 0.406 & 0\% & 0.381 & 11\% \\
fixed-3    & \underline{0.507} & 67\% & 8\% & 42\% & 9/33 & 0.496 & 2\% & 0.381 & 17\% \\
fixed-5     & 0.484 & 63\% & 7\% & 39\% & 8/33 & 0.479 & 2\% & 0.363 & 18\% \\
fixed-9    & 0.498 & 89\% & 13\% & 57\% & 16/33 & 0.433 & 5\% & 0.361 & 23\% \\
fixed-33   & 0.425 & 32\% & 1\% & 18\% & 3/33 & 0.338 & 0\% & 0.362 & 6\%  \\
\midrule
\multicolumn{10}{@{}l}{\emph{CHART: rotating window of $N$ harnesses}} \\
CHART-1    & 0.468 & 100\% & 34\% & 72\% & 22/33 & 0.392 & 71\% & \underline{0.429} & 52\% \\
CHART-3    & \textbf{0.512} & 100\% & 65\% & 85\% & 23/33 & \textbf{0.513} & 89\% & \textbf{0.437} & 59\% \\
CHART-5    & \textbf{0.512} & 100\% & 91\% & 96\% & 29/33 & \underline{0.506} & 99\% & 0.426 & 81\% \\
CHART-9    & 0.480 & 62\% & 7\% & 39\% & 7/33 & 0.414 & 3\% & 0.366 & 25\% \\
\bottomrule
\end{tabular*}
\end{table}

\subsection{Fixed Pools vs.\ Rotation}
\label{sec:motivation}

\paragraph{Pool size does not buy parallelism.} We first show the contrast-exhaustion
of \S\ref{sec:method} is not a corner case of one badly-sized pool but the outcome of
\emph{every} fixed pool. The five fixed models of Table~\ref{tab:main} differ only in
pool size, all seeded from the same \emph{learnable} mid-band harnesses at
$\sim$40--80\% base parallel rate (\S\ref{sec:method}). The cue-vs-cue-free gap is the
diagnostic: a model that has \emph{internalized} parallel search scores high on
\emph{both}, while one that merely follows instructions scores high on cue and low on
cue-free. No fixed pool lifts cue-free parallelism past $13\%$ (5\% at base) or covers
more than $16$ of the 33 harnesses, at any pool size---against $91\%$ and $29/33$ for a
rotating window. The two ends fail for opposite
reasons (\S\ref{sec:chart}): \mbox{fixed-1} consolidates on the one harness it trains
(its \parpct{} $63\%\to91\%$) and spreads to little else (cue 45\%, cue-free 2\%,
$6/33$), while \mbox{fixed-33} spreads so thin that no harness consolidates, ending
\emph{below} base on both mean \parpct{} (18\% vs.\ 24\%) and cue \parpct{} (32\% vs.\
37\%). Pool size has an interior optimum: \mbox{fixed-3} through \mbox{fixed-9}
outscore both ends on \passat{1} (0.484--0.507, against 0.456 for \mbox{fixed-1} and
0.425 for \mbox{fixed-33})---yet every fixed pool stays cue-only (cue 45--89\% against
cue-free 2--13\%).

\paragraph{Rotation improves on fixed pools at the right width.} Moving the window is
worth $+32$pp cue-free \parpct{} and $16$ more covered harnesses at $N{=}1$, $+57$pp and
$14$ more at $N{=}3$, and $+84$pp and $21$ more at $N{=}5$. Width sets a ceiling, though,
since each step's 512 rollouts split $N$ ways---$\sim$170 per harness at $N{=}3$ but only
$\sim$57 at $N{=}9$, too few to consolidate before graduation: just $4$ harnesses graduate
in the whole run and CHART-9 ends \emph{below} fixed-9 ($7\%$ vs.\ $13\%$ cue-free, $7/33$
vs.\ $16/33$). Counterintuitively, more parallelism does not keep paying either: CHART-5 reaches $91\%$ cue-free
\parpct{}, $26$pp above $N{=}3$, and still only ties it on in-pool \passat{1} ($0.512$).
We take $N{=}3$---the narrower window at equal accuracy---as the operating point for the
rest of the paper, and read these results as one hypothesis: \emph{what internalizes
parallel search is a pool that keeps turning over---enough for the behavior to
spread---rather than one that is merely large.}

% ============ MAIN RESULTS ============
\subsection{Robustness and OOD Transfer}
\label{sec:main}

\paragraph{Setup.} We now ask whether the behavior is \emph{robust, transferable, and
free of accuracy cost}: does it carry to cue-free held-out harnesses, and to a different
QA distribution answered through a different search environment (the \textbf{held-out}
and \textbf{shopping} columns of Table~\ref{tab:main})?

\paragraph{Robustness across harnesses.} Training on a fixed pool does not transfer at
all: every fixed pool sits at 0--5\% held-out \parpct{} against the untrained base's
2\%, even though every held-out harness states the per-turn call budget outright---the
behavior is tied to the trained wording, not to the opportunity. Every rotating window
from $N{=}1$ to $N{=}5$ instead reaches 71--99\%, and CHART-3 and CHART-5 take the top
two spots on held-out \passat{1} ($0.513$ and $0.506$); since CHART-3 also ties for the
best in-pool figure, the same window tops both evaluations, with no width trade-off
between accuracy and robustness. The benefit of parallelism does saturate, though:
held-out \passat{1} climbs steeply with \parpct{} at first ($0.392$ at CHART-1's 71\%,
$0.513$ at CHART-3's 89\%), but the last 10pp up to CHART-5's 99\% buys nothing
($0.506$)---the same plateau \S\ref{sec:motivation} found in-pool.

\paragraph{OOD transfer (shopping).} The shopping task changes the questions, the search
environment, and the harnesses at once, so nothing here was seen in training. Fixed pools
transfer none of the behavior to it: they parallelize on 6--23\% of turns against the
untrained base's 5\%, and their \passat{1} ($0.361$--$0.381$) straddles the base's
$0.375$, with three of the five landing below it---at this distance, training on a fixed
pool is worth nothing. Every rotating window up to $N{=}5$ instead parallelizes on
52--81\% of turns and scores $0.426$--$0.437$, beating the best fixed pool by $5.6$pp.
Since neither the task, the environment, nor the harnesses were trained on, the gap cannot
be attributed to overfitting.

\begin{table}[t]
\centering
\small
\setlength{\tabcolsep}{4pt}
\caption{CHART-3 trained 60 steps past the shared budget (200$\to$260), shown as
start$\to$end with the gain in parentheses (pp); \passat{1} omits the leading zero
(.437\,$=$\,0.437). \passat{1} rises most where \parpct{} had the most room to grow.}
\label{tab:headroom}
\begin{tabular}{@{}lcc@{}}
\toprule
task & \parpct{} 200$\to$260 & \passat{1} 200$\to$260 \\
\midrule
shopping (OOD) & 59$\to$100 ($+41$) & .437$\to$.491 ($+5.4$) \\
in-pool        & 85$\to$100 ($+15$) & .512$\to$.534 ($+2.2$) \\
held-out       & 89$\to$100 ($+11$) & .513$\to$.527 ($+1.4$) \\
\bottomrule
\end{tabular}
\end{table}
\paragraph{Beyond the shared budget.}
We give CHART-3 60 steps past the shared budget (200$\to$260) to see where the remaining
gains land. They land where parallelism still had room to grow
(Table~\ref{tab:headroom}): in-pool and held-out were already at 85--89\% \parpct{} and
move little, whereas shopping, still only 59\% parallel, rises to 100\% and gains
$5.4$pp \passat{1}---the largest gain of the three, on the one task the model never
trained on. We use this checkpoint as the CHART substrate in \S\ref{sec:metasearch}.

% ============ META-HARNESS SEARCH PAYOFF ============
\subsection{Meta-Harness Search}
\label{sec:metasearch}

\paragraph{Setup.} We ask whether a more-trained substrate lets the \emph{same}
meta-harness search \citep{lee2026metaharness} find a better harness. For each of three
frozen checkpoints---the untrained \textbf{base}, \textbf{CHART-1}, and
\textbf{CHART-3}---we run an identical search that tunes only the system
prompt, never the weights: from a hand-written seed pool we keep the top 5 by
\emph{validation} pass@1, then each round every seed proposes 5 mutations
($5\times5{=}25$ candidates), we re-rank the \emph{entire cumulative history}, and
the top 5 reseed the next round. We score the final top 5 \emph{once} on a held-out
\emph{test} set and report \emph{best-of-search}, the validation champion's test
score, and \emph{top-5 mean}, their mean. We search two tasks: the \textbf{shopping
agent task}, using the two splits of the shopping product-search set
(\S\ref{sec:setup}), and \textbf{deep-research} QA, with the \texttt{miroverse}
validation set of \S\ref{sec:setup} and a test set of FRAMES
\citep{krishna2024frames} and seal\_hard \citep{pham2025sealqa}.\footnote{200
questions randomly sampled from FRAMES (compute limits) plus 254 from seal\_hard.}

\begin{table}[t]
\centering
\caption{Meta-harness search over three training levels: a more-trained substrate searches to a better harness on both tasks and
metrics. deep-research pass@1 and \parpct{} are computed over the pooled
FRAMES\,$+$\,seal\_hard instances ($200{+}254$);
per-set breakout in Appendix~\ref{app:metasearch}.}
\label{tab:metasearch}
\begin{tabular}{@{}llcccc@{}}
\toprule
& & \multicolumn{2}{c}{pass@1} & \multicolumn{2}{c}{\parpct} \\
\cmidrule(lr){3-4}\cmidrule(lr){5-6}
Task & Model & best-of-search & top-5 mean & best-of-search & top-5 mean \\
\midrule
\multirow{3}{*}{shopping}
 & base            & 0.415 & 0.451 & 100\% & 99\% \\
 & CHART-1         & 0.445 & 0.464 & 100\% & 100\% \\
 & \textbf{CHART-3} & \textbf{0.515} & \textbf{0.485} & 100\% & 100\% \\
\midrule
\multirow{3}{*}{deep-research}
 & base            & 0.409 & 0.387 & 6\%  & 19\% \\
 & CHART-1         & 0.473 & 0.477 & 90\% & 91\% \\
 & \textbf{CHART-3} & \textbf{0.535} & \textbf{0.533} & 100\% & 100\% \\
\bottomrule
\end{tabular}
\end{table}

\paragraph{Search outcome.}
\emph{Any} training beats no training (Table~\ref{tab:metasearch}): base is weakest on
both tasks ($0.415$ best-of-search on shopping, $0.409$ on deep-research), and both
rotation checkpoints search to a clearly higher outcome. More training makes a better
substrate still: \textbf{CHART-3} is strongest on every metric, reaching
$0.515$ and $0.535$ best-of-search---$+10.0$ and $+12.6$pp over base---with CHART-1
between. The base $<$ CHART-1 $<$ CHART-3 ordering is monotone and holds for the top-5 mean as
well as the single winner, ruling out a lucky harness under noisy $n{=}1$ evaluation.
The \parpct{} columns show where the deep-research gain comes from: \parpct{}
generalizes $6\%\to90\%\to100\%$ and pass@1 rises with it, a $31\%$ relative
improvement over base. Per-task mechanisms in Appendix~\ref{app:metasearch}.

\paragraph{Model $\times$ harness cross.}
Since the search above scores each model under its own search-tuned harness---conflating a
stronger model with a harness merely tuned to flatter it---we cross the three models
with their best-of-search champions in a $3\times3$ grid on the shopping test set,
sampling each cell $n{=}8$ times to suppress the $n{=}1$ noise of the search above
(Table~\ref{tab:cross3}). Each \textbf{row} scores one model across the three prompts:
its \textbf{mean} is harness-independent quality, its \textbf{spread} (max$-$min) its
harness-sensitivity.

\begin{table}[t]
\centering
\caption{$3\times3$ training $\times$ harness cross on the shopping test set,
$n{=}8$ samples per cell: more training both helps---row mean rises, every column monotone
base $<$ CHART-1 $<$ CHART-3---and flattens harness-sensitivity, as row spread shrinks.
Best per column in bold.}
\label{tab:cross3}
\begin{tabular}{@{}lccccc@{}}
\toprule
& \multicolumn{3}{c}{harness (each model's val champion)} & & \\
\cmidrule(lr){2-4}
Model & base champ & CHART-1 champ & CHART-3 champ & row mean & row spread \\
\midrule
base            & 0.418 & 0.316 & 0.347 & 0.360 & 0.102 \\
CHART-1         & 0.449 & 0.429 & 0.431 & 0.436 & \textbf{0.020} \\
\textbf{CHART-3} & \textbf{0.490} & \textbf{0.460} & \textbf{0.450} & \textbf{0.467} & 0.040 \\
\bottomrule
\end{tabular}
\end{table}

The row mean rises monotonically from $0.360$ at base to $0.436$ at CHART-1 to $0.467$
at CHART-3 ($+10.7$pp; the last $+3.1$pp from widening the window $1\to3$). Because the
mean is over a \emph{common} harness set, it reflects the model itself, settling that
confound. Meanwhile the spread shrinks from $10.2$pp at base to
$2$--$4$pp once trained: the base model craters to $0.316$ under the CHART-1 champion
prompt that pushes a parallelism it cannot yet execute, whereas trained models score
nearly the same under any prompt---both better and flatter over \emph{harnesses}.

% ============ ABLATIONS ============
\subsection{Ablations}
\label{sec:ablation}
We isolate three design choices behind rotation---its \emph{length}, its \emph{refill
rule}, and its \emph{graduation trigger}---each run varying only that knob.

\begin{table}[t]
\centering
\small
\setlength{\tabcolsep}{4pt}
\setlength{\belowcaptionskip}{6pt}
\begin{minipage}[t]{0.46\linewidth}
\centering
\caption{Pool frozen at the fork vs.\ rotating throughout. In parentheses, how many
covered harnesses are cue-free; \passat{1} averages the three tasks.}
\label{tab:continued}
\begin{tabular}{@{}lcc@{}}
\toprule
step 200 & early-stop & full-run \\
\midrule
coverage (cue-free) & 14 (1) & \textbf{23 (4)} \\
\addlinespace[2pt]
\multicolumn{3}{@{}l}{\parpct} \\
\quad cue (A/C)      & 85\% & \textbf{100\%} \\
\quad cue-free (B/D) & 14\% & \textbf{65\%}  \\
\quad held-out       & 7\%  & \textbf{89\%}  \\
\quad shopping (OOD) & 28\% & \textbf{59\%}  \\
\addlinespace[2pt]
mean \passat{1}     & .461   & \textbf{.487} \\
\bottomrule
\end{tabular}
\end{minipage}\hfill
\begin{minipage}[t]{0.50\linewidth}
\centering
\caption{Refill of a slot freed by graduation, at $N{=}3$ from the same seed.}
\label{tab:refill}
\begin{tabular}{@{}lccc@{}}
\toprule
refill & \mbox{fixed-3} & random & CHART-3 \\
\midrule
\multicolumn{4}{@{}l}{\passat{1}} \\
\quad in-pool  & .507 & .372 & \textbf{.512} \\
\quad held-out & .496 & .306 & \textbf{.513} \\
\quad shopping & .381 & .333 & \textbf{.437} \\
\addlinespace[2pt]
\multicolumn{4}{@{}l}{\parpct} \\
\quad in-pool  & 42\% & 46\% & \textbf{85\%} \\
\quad held-out & 2\%  & 8\%  & \textbf{89\%} \\
\quad shopping & 17\% & 18\% & \textbf{59\%} \\
\bottomrule
\end{tabular}
\end{minipage}
\end{table}
\paragraph{Rotation length.}
The first knob, with $N{=}3$ fixed, is how long we rotate: is rotation only a
\emph{warm-up}, switchable off once the model has a general disposition to parallelize?
Two runs share their first 100 steps, then fork---\textbf{early-stop} freezes the pool
and keeps training on it, \textbf{full-run} keeps rotating. It is not: passive
generalization stalls at the cue boundary. Early-stop does keep generalizing on its own,
but only to cued styles: it covers 14 harnesses against full-run's 23, and just $1$ of
them is cue-free against $4$. The parallel rates of Table~\ref{tab:continued} put the
same boundary in plain view: the two are close wherever the prompt asks for parallel
search and far apart everywhere it does not.
Rotating longer is not costly either: full-run also ends with the higher mean
\passat{1}. Rotation must therefore run throughout, not only early.

\paragraph{Refill rule.}
The second knob is \emph{which} harness fills a freed slot: we replace CHART's
learnable-band draw ($\parpct\in[0.40,0.80]$, \S\ref{sec:method}) with a uniform draw over
the not-yet-graduated harnesses (Table~\ref{tab:refill}). Random refill is the weakest arm in
the study, losing to CHART-3 and even to \mbox{fixed-3}---to not rotating at all---by $12$pp
of \passat{1} averaged over the three tasks. Its draws land on near-zero-\parpct{} harnesses---nothing
to consolidate---and by step $180$ no seed harness is left. The cost is accuracy, not
parallel usage: \parpct{} holds at a fixed pool's level while \passat{1} drops below
it---rotation is only as good as what it rotates \emph{in}.

\paragraph{Graduation trigger.}
The third knob is \emph{when} a harness leaves once its parallel rate reaches the
graduation bar. \textbf{Eval-only} checks only every $\sim$20 steps, so a passing
harness overshoots to $\parpct\approx0.95$--$1.0$ before leaving; \textbf{EMA+eval} also
tracks a per-step exponential moving average (EMA) of each harness's parallel rate over
its \emph{training} rollouts and graduates the instant that estimate crosses the bar.
Table~\ref{tab:trigger} shows faster graduation loses on
every metric: both reach high cue \parpct{} (100\% vs.\ 82\%), but on cue-free
eval-only holds 65\% while EMA collapses to 13\%, and held-out pass@1 follows at
0.513 vs.\ 0.372. EMA also graduates \emph{fewer} harnesses despite leaving each
sooner---4 vs.\ 12, stalling by step 60 (Table~\ref{tab:timeline}); its active
harnesses stay stuck at $0.54$--$0.63$, never nearing the bar. That overshoot is where
the behavior \emph{consolidates} into a default: EMA skips it, so parallel never
generalizes and the pool stops turning over.

\begin{table}[!ht]
\centering
\small
\begin{minipage}[t]{0.56\linewidth}
\centering
\caption{Graduation trigger: eval-only vs.\ EMA+eval at step 200.}
\label{tab:trigger}
\begin{tabular}{@{}lcc@{}}
\toprule
Metric  & eval-only & EMA+eval \\
\midrule
harnesses graduated              & 12   & 4    \\
cue (A/C) \parpct{}              & 100\% & 82\% \\
cue-free (B/D) \parpct{}          & 65\% & 13\% \\
held-out pass@1                  & 0.513 & 0.372 \\
held-out \parpct{}               & 89\%  & 7\%  \\
\bottomrule
\end{tabular}
\end{minipage}\hfill
\begin{minipage}[t]{0.40\linewidth}
\centering
\caption{Cumulative harnesses graduated by step: EMA stalls after step 60 while
eval-only keeps graduating.}
\label{tab:timeline}
\begin{tabular}{@{}l@{\ }ccccc@{}}
\toprule
step & 20 & 60 & 100 & 140 & 200 \\
\midrule
EMA+eval  & 2 & 4 & 4 & 4  & 4  \\
eval-only & 0 & 2 & 4 & 7  & 12 \\
\bottomrule
\end{tabular}
\end{minipage}
\end{table}

% ============ CONCLUSION ============
\section{Discussion and Limitations}
\label{sec:conclusion}

We showed that RL-learned agent behavior can be harness-local---an agent trained to
search in parallel reverts to serial search when the harness is reworded---and that this
is a training-dynamics failure: loss of within-group contrast after saturation.
\textbf{CHART}, a curriculum over the harness, keeps that contrast alive and spreads
harness-robust parallel behavior to unseen styles, new QA distributions, and new search
environments at no accuracy cost, which in turn flattens the harness landscape and makes
downstream meta-harness search more effective.

\paragraph{Limitations.} We study one behavior---intra-turn parallel search---on one
base model, Qwen3-30B-A3B, so generalization to other agent behaviors and model families
is open. We also vary only the harness's system prompt, leaving tool-interface changes
such as renamed or distractor tools to future work.

\bibliographystyle{plainnat}
\bibliography{references}

\begin{thebibliography}{35}
\providecommand{\natexlab}[1]{#1}
\providecommand{\url}[1]{\texttt{#1}}
\expandafter\ifx\csname urlstyle\endcsname\relax
  \providecommand{\doi}[1]{doi: #1}\else
  \providecommand{\doi}{doi: \begingroup \urlstyle{rm}\Url}\fi

\bibitem[Agrawal et~al.(2025)Agrawal, Tan, Soylu, Ziems, Khare, Opsahl-Ong,
  Singhvi, Shandilya, Ryan, Jiang, Potts, Sen, Dimakis, Stoica, Klein, Zaharia,
  and Khattab]{agrawal2025gepa}
Lakshya~A. Agrawal, Shangyin Tan, Dilara Soylu, Noah Ziems, Rishi Khare, Krista
  Opsahl-Ong, Arnav Singhvi, Herumb Shandilya, Michael~J. Ryan, Meng Jiang,
  Christopher Potts, Koushik Sen, Alexandros~G. Dimakis, Ion Stoica, Dan Klein,
  Matei Zaharia, and Omar Khattab.
\newblock Gepa: Reflective prompt evolution can outperform reinforcement
  learning.
\newblock \emph{arXiv preprint arXiv:2507.19457}, 2025.

\bibitem[Anthropic(2025)]{anthropic2025claudecode}
Anthropic.
\newblock Claude code.
\newblock \url{https://claude.com/product/claude-code}, 2025.

\bibitem[Bai and Shi(2026)]{maspromptbench2026}
Juyang Bai and Laixi Shi.
\newblock Mas-promptbench: When does prompt optimization improve multi-agent
  llm systems?
\newblock \emph{arXiv preprint arXiv:2606.23664}, 2026.

\bibitem[Bengio et~al.(2009)Bengio, Louradour, Collobert, and
  Weston]{bengio2009curriculum}
Yoshua Bengio, J{\'e}r{\^o}me Louradour, Ronan Collobert, and Jason Weston.
\newblock Curriculum learning.
\newblock In \emph{Proceedings of the 26th International Conference on Machine
  Learning (ICML)}, 2009.

\bibitem[Guo et~al.(2023)Guo, Wang, Guo, Li, Song, Tan, Liu, Bian, and
  Yang]{guo2023evoprompt}
Qingyan Guo, Rui Wang, Junliang Guo, Bei Li, Kaitao Song, Xu~Tan, Guoqing Liu,
  Jiang Bian, and Yujiu Yang.
\newblock Evoprompt: Connecting llms with evolutionary algorithms yields
  powerful prompt optimizers.
\newblock \emph{arXiv preprint arXiv:2309.08532}, 2023.

\bibitem[Jiang et~al.(2025)]{jiang2025vcrl}
Guochao Jiang et~al.
\newblock Vcrl: Variance-based curriculum reinforcement learning for large
  language models.
\newblock \emph{arXiv preprint arXiv:2509.19803}, 2025.

\bibitem[Jin et~al.(2025)]{jin2025searchr1}
Bowen Jin et~al.
\newblock Search-r1: Training llms to reason and leverage search engines with
  reinforcement learning.
\newblock \emph{arXiv preprint arXiv:2503.09516}, 2025.

\bibitem[Kalyan and Andrews(2025)]{kalyan2025longhorizon}
Vivek Kalyan and Martin Andrews.
\newblock Reinforcement learning for long-horizon multi-turn search agents.
\newblock \emph{arXiv preprint arXiv:2510.24126}, 2025.

\bibitem[Kim et~al.(2023)Kim, Moon, Tabrizi, Lee, Mahoney, Keutzer, and
  Gholami]{kim2023llmcompiler}
Sehoon Kim, Suhong Moon, Ryan Tabrizi, Nicholas Lee, Michael~W. Mahoney, Kurt
  Keutzer, and Amir Gholami.
\newblock An llm compiler for parallel function calling.
\newblock \emph{arXiv preprint arXiv:2312.04511}, 2023.

\bibitem[{Kimi Team}(2026)]{kimiteam2026k3}
{Kimi Team}.
\newblock {Kimi K3: Open Frontier Intelligence}.
\newblock \emph{arXiv preprint arXiv:2607.24653}, 2026.

\bibitem[Krishna et~al.(2024)]{krishna2024frames}
Satyapriya Krishna et~al.
\newblock {Fact, Fetch, and Reason: A Unified Evaluation of Retrieval-Augmented
  Generation}.
\newblock \emph{arXiv preprint arXiv:2409.12941}, 2024.

\bibitem[Lee et~al.(2026)Lee, Nair, Zhang, Lee, Khattab, and
  Finn]{lee2026metaharness}
Yoonho Lee, Roshen Nair, Qizheng Zhang, Kangwook Lee, Omar Khattab, and Chelsea
  Finn.
\newblock {Meta-Harness: End-to-End Optimization of Model Harnesses}.
\newblock \emph{arXiv preprint arXiv:2603.28052}, 2026.

\bibitem[Li et~al.(2025)Li, Zou, and Liu]{li2025torl}
Xuefeng Li, Haoyang Zou, and Pengfei Liu.
\newblock Torl: Scaling tool-integrated rl.
\newblock \emph{arXiv preprint arXiv:2503.23383}, 2025.

\bibitem[Liang et~al.(2025)Liang, Li, Gong, et~al.]{liang2025sws}
Xiao Liang, Zhong-Zhi Li, Yeyun Gong, et~al.
\newblock Sws: Self-aware weakness-driven problem synthesis in reinforcement
  learning for llm reasoning.
\newblock \emph{arXiv preprint arXiv:2506.08989}, 2025.

\bibitem[Lu et~al.(2022)Lu, Bartolo, Moore, Riedel, and
  Stenetorp]{lu2022fantastically}
Yao Lu, Max Bartolo, Alastair Moore, Sebastian Riedel, and Pontus Stenetorp.
\newblock Fantastically ordered prompts and where to find them: Overcoming
  few-shot prompt order sensitivity.
\newblock \emph{arXiv preprint arXiv:2104.08786}, 2022.

\bibitem[{MiroMind AI}(2025)]{miroverse2025}
{MiroMind AI}.
\newblock {MiroVerse-v0.1}.
\newblock \url{https://huggingface.co/datasets/miromind-ai/MiroVerse-v0.1},
  2025.
\newblock HuggingFace dataset.

\bibitem[Mizrahi et~al.(2024)Mizrahi, Kaplan, Malkin, Dror, Shahaf, and
  Stanovsky]{mizrahi2024state}
Moran Mizrahi, Guy Kaplan, Dan Malkin, Rotem Dror, Dafna Shahaf, and Gabriel
  Stanovsky.
\newblock State of what art? a call for multi-prompt llm evaluation.
\newblock \emph{arXiv preprint arXiv:2401.00595}, 2024.

\bibitem[Panaganti et~al.(2026)Panaganti, Liang, Yu, et~al.]{panaganti2026gdro}
Kishan Panaganti, Zhenwen Liang, Wenhao Yu, et~al.
\newblock Group distributionally robust optimization-driven reinforcement
  learning for llm reasoning.
\newblock \emph{arXiv preprint arXiv:2601.19280}, 2026.

\bibitem[Pham et~al.(2025)Pham, Nguyen, Zunjare, Chen, Tseng, and
  Vu]{pham2025sealqa}
Thinh Pham, Nguyen Nguyen, Pratibha Zunjare, Weiyuan Chen, Yu-Min Tseng, and
  Tu~Vu.
\newblock {SealQA: Raising the Bar for Reasoning in Search-Augmented Language
  Models}.
\newblock \emph{arXiv preprint arXiv:2506.01062}, 2025.

\bibitem[Qi et~al.(2024)Qi, Liu, Iong, Lai, Sun, Yang, Sun, Yang, Yao, Zhang,
  et~al.]{qi2024webrl}
Zehan Qi, Xiao Liu, Iat~Long Iong, Hanyu Lai, Xueqiao Sun, Xinyue Yang, Jiadai
  Sun, Yu~Yang, Shuntian Yao, Tianjie Zhang, et~al.
\newblock Webrl: Training llm web agents via self-evolving online curriculum
  reinforcement learning.
\newblock \emph{arXiv preprint arXiv:2411.02337}, 2024.

\bibitem[Romanou et~al.(2026)Romanou, Ibrahim, Ross, Shaib, Oktar, Bell,
  Ovalle, Dodge, Bosselut, Sinha, and Williams]{brittlebench2026}
Angelika Romanou, Mark Ibrahim, Candace Ross, Chantal Shaib, Kerem Oktar,
  Samuel~J. Bell, Anaelia Ovalle, Jesse Dodge, Antoine Bosselut, Koustuv Sinha,
  and Adina Williams.
\newblock Brittlebench: Quantifying llm robustness via prompt sensitivity.
\newblock \emph{arXiv preprint arXiv:2603.13285}, 2026.

\bibitem[Sanh et~al.(2021)Sanh, Webson, Raffel, Bach, Sutawika,
  et~al.]{sanh2021t0}
Victor Sanh, Albert Webson, Colin Raffel, Stephen~H. Bach, Lintang Sutawika,
  et~al.
\newblock Multitask prompted training enables zero-shot task generalization.
\newblock \emph{arXiv preprint arXiv:2110.08207}, 2021.

\bibitem[Sclar et~al.(2023)Sclar, Choi, Tsvetkov, and
  Suhr]{sclar2023quantifying}
Melanie Sclar, Yejin Choi, Yulia Tsvetkov, and Alane Suhr.
\newblock Quantifying language models' sensitivity to spurious features in
  prompt design or: How i learned to start worrying about prompt formatting.
\newblock \emph{arXiv preprint arXiv:2310.11324}, 2023.

\bibitem[Shao et~al.(2024)Shao, Wang, Zhu, Xu, Song, Bi,
  et~al.]{shao2024deepseekmath}
Zhihong Shao, Peiyi Wang, Qihao Zhu, Runxin Xu, Junxiao Song, Xiao Bi, et~al.
\newblock Deepseekmath: Pushing the limits of mathematical reasoning in open
  language models.
\newblock \emph{arXiv preprint arXiv:2402.03300}, 2024.

\bibitem[Song et~al.(2025)]{song2025r1searcher}
Huatong Song et~al.
\newblock R1-searcher: Incentivizing the search capability in llms via
  reinforcement learning.
\newblock \emph{arXiv preprint arXiv:2503.05592}, 2025.

\bibitem[Sun et~al.(2025)]{sun2025zerosearch}
Hao Sun et~al.
\newblock Zerosearch: Incentivize the search capability of llms without
  searching.
\newblock \emph{arXiv preprint arXiv:2505.04588}, 2025.

\bibitem[Sun et~al.(2023)Sun, Shaib, and Wallace]{sun2023robustness}
Jiuding Sun, Chantal Shaib, and Byron~C. Wallace.
\newblock Evaluating the zero-shot robustness of instruction-tuned language
  models.
\newblock \emph{arXiv preprint arXiv:2306.11270}, 2023.

\bibitem[Wei et~al.(2021)Wei, Bosma, Zhao, Guu, Yu, Lester, Du, Dai, and
  Le]{wei2021flan}
Jason Wei, Maarten Bosma, Vincent~Y. Zhao, Kelvin Guu, Adams~Wei Yu, Brian
  Lester, Nan Du, Andrew~M. Dai, and Quoc~V. Le.
\newblock Finetuned language models are zero-shot learners.
\newblock \emph{arXiv preprint arXiv:2109.01652}, 2021.

\bibitem[Xi et~al.(2024)Xi, Chen, Hong, et~al.]{xi2024reverse}
Zhiheng Xi, Wenxiang Chen, Boyang Hong, et~al.
\newblock Training large language models for reasoning through reverse
  curriculum reinforcement learning.
\newblock \emph{arXiv preprint arXiv:2402.05808}, 2024.

\bibitem[Yang et~al.(2025)Yang, Li, Yang, Zhang, et~al.]{qwen3}
An~Yang, Anfeng Li, Baosong Yang, Beichen Zhang, et~al.
\newblock Qwen3 technical report.
\newblock \emph{arXiv preprint arXiv:2505.09388}, 2025.

\bibitem[Yang et~al.(2023)Yang, Wang, Lu, et~al.]{yang2023opro}
Chengrun Yang, Xuezhi Wang, Yifeng Lu, et~al.
\newblock Large language models as optimizers.
\newblock \emph{arXiv preprint arXiv:2309.03409}, 2023.

\bibitem[Young et~al.(2025)Young, Gillins, and Matthews]{young2025follow}
Richard~J. Young, Brandon Gillins, and Alice~M. Matthews.
\newblock When models can't follow: Testing instruction adherence across 256
  llms.
\newblock \emph{arXiv preprint arXiv:2510.18892}, 2025.

\bibitem[Zhang et~al.(2024)Zhang, Yuan, and Avestimehr]{zhang2024revisiting}
Tuo Zhang, Jinyue Yuan, and Salman Avestimehr.
\newblock Revisiting opro: The limitations of small-scale llms as optimizers.
\newblock \emph{arXiv preprint arXiv:2405.10276}, 2024.

\bibitem[Zhou et~al.(2022)Zhou, Muresanu, Han, Paster, Pitis, Chan, and
  Ba]{zhou2022ape}
Yongchao Zhou, Andrei~Ioan Muresanu, Ziwen Han, Keiran Paster, Silviu Pitis,
  Harris Chan, and Jimmy Ba.
\newblock Large language models are human-level prompt engineers.
\newblock \emph{arXiv preprint arXiv:2211.01910}, 2022.

\bibitem[Zhu et~al.(2023)Zhu, Wang, Zhou, Wang, Chen, Wang, Yang, Ye, Zhang,
  Gong, and Xie]{zhu2023promptrobust}
Kaijie Zhu, Jindong Wang, Jiaheng Zhou, Zichen Wang, Hao Chen, Yidong Wang,
  Linyi Yang, Wei Ye, Yue Zhang, Neil~Zhenqiang Gong, and Xing Xie.
\newblock Promptrobust: Towards evaluating the robustness of large language
  models on adversarial prompts.
\newblock \emph{arXiv preprint arXiv:2306.04528}, 2023.

\end{thebibliography}

\appendix
% ============ APPENDIX ============
\section{Harness Pool Design}
\label{app:harnesses}
The 33-harness pool was authored with Claude Code and
organized into four groups by how much the harness tells the model what to do
(\S\ref{sec:setup}). \textbf{Group A (Explicit, 10)} states the parallel strategy
outright and serves as the training anchor---the model already parallelizes here,
so it gives stable positive signal; variants cover wording, persona (``fact
checker,'' ``academic researcher''), verbosity, output formatting, and language
(including Chinese-language variants that state the per-turn parallel budget in
Chinese, e.g.\ ``5 parallel calls per turn''). \textbf{Group B
(Implicit, 9)} hints at the behavior without naming it (``maximize information
per round,'' ``multiple sources,'' ``batch your queries,'' time pressure, and a
Chinese implicit variant). \textbf{Group C (Framework, 9)} recasts the same task
in popular agent formats---thought/action/observation (ReAct-style),
plan-and-execute, an explicit agent loop, hypothesis-driven investigation,
decompose-and-search, iterative refinement, a terse machine-readable system spec,
a conversational framing, and a bare toolbox listing---to test
format-invariance. \textbf{Group D (Minimal, 5)} strips behavioral cues to near
zero (a two-line minimal prompt, a tools-only listing, a one-sentence
instruction, the task buried in unrelated boilerplate, and a long tool list with
no strategy guidance), testing whether parallel search has been internalized
rather than instructed. One further adversarial prompt was authored---it actively
steers toward serial, one-query-at-a-time search---but is excluded from the pool
and from every eval reported here, so the pool is $33 = 10+9+9+5$.

\paragraph{Representative excerpts.} We show one prompt per group rather than all
33 templates. All four expose the same two tools (\texttt{ngs\_search},
\texttt{scrape}), require the Hermes \texttt{<tool\_call>} format, and ask for the
answer in \verb|\boxed{}|; they differ only in how much strategy they prescribe.
Placeholders \texttt{\{max\_turns\}}, \texttt{\{max\_parallel\}}, and
\texttt{\{total\_tool\_calls\}} are filled in at rollout time from the turn
curriculum.

\emph{Group A (explicit)}---the parallel budget and the ``execute in parallel''
instruction are stated directly:
{\small\begin{verbatim}
You are a research assistant that finds precise answers to
questions using web search.
...
## Efficiency Rules
- You have **{max_turns} turns** with up to **{max_parallel}
  parallel calls per turn** ({total_tool_calls} total calls).
  Exhausting all turns without answering discards your response.
- If a search returns the answer clearly in a snippet, stop and
  answer immediately. Do not search further to "verify".
- If 2 consecutive searches fail, STOP rephrasing the same query.
  Decompose the question differently, search for a related
  entity, or try a completely different angle.

## Tool Call Format
Use the Hermes tool call format. You can issue multiple calls in
one response -- they execute in parallel:
<tool_call>
{"name": "ngs_search", "arguments": {"query": "first query"}}
</tool_call>
<tool_call>
{"name": "ngs_search", "arguments": {"query": "second query"}}
</tool_call>
Use parallel calls for independent queries; use sequential calls
when a query depends on a previous result.
\end{verbatim}}

\emph{Group B (implicit)}---no mention of parallelism or of a per-turn call
budget; only an efficiency objective the model may satisfy any way it likes:
{\small\begin{verbatim}
## Guidelines
- You have {max_turns} turns to find the answer.
- Maximize information gathered per round of tool use.
- If a search returns the answer in a snippet, stop and answer
  immediately.
- If searches fail, try a completely different angle.

## Tool Call Format
<tool_call>
{"name": "ngs_search", "arguments": {"query": "your query"}}
</tool_call>
\end{verbatim}}

\emph{Group C (framework)}---the task is re-expressed in a third-party agent
idiom; concurrency is mentioned only as a property of the framework's action
step:
{\small\begin{verbatim}
You are an agent that answers questions through iterative
research.

At each step, follow the Thought -> Action -> Observation cycle:
- Thought: What do I need to find? What's my search strategy?
- Action: Issue tool calls to search or read pages.
- Observation: Analyze the results.

## Constraints
- Maximum {max_turns} action rounds, up to {max_parallel}
  actions per round
- Total action budget: {total_tool_calls}

You may take multiple actions per thought step -- they execute
simultaneously.
\end{verbatim}}

\emph{Group D (minimal)}---the entire harness, with no strategy, persona,
budget, or behavioral cue of any kind:
{\small\begin{verbatim}
Search the web to answer the question.

Tools: ngs_search(query, num_results),
       scrape(url, instruction).

Format:
<tool_call>
{"name": "...", "arguments": {...}}
</tool_call>

Answer: \boxed{answer}
\end{verbatim}}

\section{Budget-Matched Parallel vs.\ Sequential Search}
\label{app:parvsseq}
We test the paper's premise---that parallel search improves both accuracy and
efficiency (\S\ref{sec:intro})---with a budget-matched pair of single-harness
runs. Both train from the same base model (Qwen3-30B-A3B) on the same data with
the same additive reward under one fixed harness, and differ \emph{only} in how
the $\sim$25-search budget is spent: the \emph{sequential} baseline issues one
call per turn (up to $T{=}25$ turns, pure serial depth), while the \emph{parallel}
harness---the fixed-1 harness used elsewhere in the paper---issues several
calls per turn. The reward adds no parallel or efficiency bonus, so any gap
reflects the search strategy alone. Both are evaluated on the same 150-prompt val set
($\passat{1}$, $N{=}1$).

\begin{table}[t]
\centering
\caption{Budget-matched single-harness comparison (deep-research val, 150 prompts,
$N{=}1$). Same base model and $\sim$25-search budget; parallel is more accurate and
far more efficient. Means are per prompt.}
\label{tab:parvsseq}
\begin{tabular}{@{}lcccccc@{}}
\toprule
Strategy & \passat{1} & turns & LLM calls & searches & input tok & output tok \\
\midrule
Sequential ($P{=}1$, $T{=}25$)          & 0.473 & 10.9 & 11.9 & 10.9 & 67.0k & 2.1k \\
Parallel (fixed-1, $P{=}5$, $T{=}5$) & \textbf{0.513} & \textbf{2.5} & \textbf{3.4} & 10.5 & \textbf{26.5k} & 2.2k \\
\bottomrule
\end{tabular}
\end{table}

Table~\ref{tab:parvsseq} shows parallel search is Pareto-better. It is more
\emph{accurate} ($+4.0$pp $\passat{1}$; the gain concentrates on the learnable
slice, $+11.4$pp, with easy saturated and hard tied within $N{=}1$ noise) and more
\emph{efficient}: issuing $\sim$4 calls per turn, it reaches the same budget in
$4.4\times$ fewer turns and $3.5\times$ fewer LLM calls. The efficiency is
structural rather than shorter generations---both runs issue about the same number
of searches ($10.5$ vs.\ $10.9$) and produce about the same output length ($2.2$k
vs.\ $2.1$k tokens); the saving is that the sequential agent re-encodes its growing
context every turn and so processes $2.5\times$ more input tokens ($67$k vs.\
$26$k). Here \emph{input tokens} counts the prompt (prefill) tokens the model
reads summed over all its LLM calls in a trajectory---since the full conversation
so far (system prompt, question, and every prior turn's reasoning and tool
results) is re-sent on each call, this grows with the number of turns---and
\emph{output tokens} counts the tokens the model generates, summed over the same
calls; both are then averaged over the 150 prompts. Input tokens thus measures
how much context the model must re-encode end-to-end, while output tokens measures
raw generation.

\section{Eval Sample-Size Validation}
\label{app:evalsize}
The graduation trigger (\S\ref{sec:method}) reads a lightweight in-pool eval of
50 datapoints per harness, so its \parpct{} estimate carries roughly $\pm8$pp of
binomial noise. We validate that this is nonetheless enough to make the same
graduation decision as the $3\times$ larger eval used for reporting. On two
checkpoints of the rotation run (iter\_59 and iter\_79) we ran the full standalone
eval---33 harnesses $\times$ 150 questions, $N{=}1$, temperature 1.0---and
asked, for each harness, whether the two evals make the same graduate /
keep-training call at the graduation threshold.

They agree on 32/33 harnesses (97\%) at both checkpoints. The single disagreement
is the same harness each time, and it is one whose parallel rate genuinely
oscillates across steps rather than one the eval mis-measures (50-datapoint eval
reads $90\%$, standalone reads $78\%$). The consequence of such a false graduation
is mild: the harness leaves the active window slightly early and is replaced by a
mid-band harness that still provides contrast signal. We therefore use
50-datapoint eval for graduation and checkpoint selection, and 150-question eval
for all reported numbers.

\section{Meta-Harness Search Breakdown}
\label{app:metasearch}

\paragraph{The lever differs by task.} In both settings, training equips the model
to exploit a harness the base model cannot, but through different mechanisms. On
\textbf{deep-research}, the base model searches serially by default---its champion
harness operates at $6\%$ \parpct{}, and forcing parallel execution on the base
model \emph{reduces} accuracy---so the behavior must be trained before search can
exploit it; the benefit of training is thus the emergence of parallelism (the
selected harness's \parpct{} increases $6\%\to90\%\to100\%$ along
base$\to$CHART-1$\to$CHART-3, accounting for the $+12.6$pp improvement). On
\textbf{shopping}, parallelism is already saturated on every checkpoint ($100\%$
\parpct{} even at base, as the winning harness prescribes fully parallel search),
yet pass@1 still increases by $10.0$pp with \parpct{} fixed at $100\%$; here
training instead enables deeper verification, as the top-5 harnesses issue more
tool calls (mean $8.7\to13.1\to13.4$, predominantly \texttt{product\_info}
attribute checks, at a stable rate of $\sim$4.7 calls per turn). In both cases the
trained substrate lets search select a stronger harness---one that parallelizes
(deep-research) or verifies more thoroughly (shopping) in a manner unavailable to
the base model.

\paragraph{Per-set test breakdown.} Table~\ref{tab:metasearch} in
\S\ref{sec:metasearch} reports the deep-research pass@1 pooled over the FRAMES and
seal\_hard instances. Table~\ref{tab:metasearch-breakout} breaks that pooled score
into its two component test sets for the validation champion (best-of-search). Both
sets rise monotonically with training, so the deep-research gain is not driven by a
single dataset.

\begin{table}[t]
\centering
\caption{Deep-research best-of-search pass@1, per test set. The two sets are
pooled at the instance level ($200{+}254$ questions) into the deep-research pass@1
row of Table~\ref{tab:metasearch}; because pooling weights by set size, the overall
row is not the simple average of the two. Both rise monotonically base
$\to$ CHART-1 $\to$ CHART-3.}
\label{tab:metasearch-breakout}
\begin{tabular}{@{}lccc@{}}
\toprule
Test set & base & CHART-1 & CHART-3 \\
\midrule
FRAMES     & 0.550 & 0.645 & \textbf{0.705} \\
seal\_hard & 0.268 & 0.350 & \textbf{0.378} \\
\midrule
overall (pooled) & 0.409 & 0.473 & \textbf{0.535} \\
\bottomrule
\end{tabular}
\end{table}

\end{document}